# On-line signature verification system with failure to enroll managing

Joan Fabregas, Marcos Faundez-Zanuy

Escola Universitaria Politecnica de Mataro

Avda. Puig i Cadafalch 101-111, 08303, MATARO (Barcelona), Spain.

*Abstract*

*In this paper we simulate a real biometric verification system based on on-line signatures. For this purpose we have split the MCYT signature database in three subsets: one for classifier training, another for system adjustment and a third one for system testing simulating enrollment and verification. This context corresponds to a real operation, where a new user tries to enroll an existing system and must be automatically guided by the system in order to detect the failure to enroll situations. The main contribution of this work is the management of failure to enroll situations by means of a new proposal, called intelligent enrollment, which consists of consistency checking in order to automatically reject low quality samples. This strategy lets to enhance the verification errors up to 22% when leaving out 8% of the users. In this situation 8% of the people cannot be enrolled in the system and must be verified by other biometrics or by human abilities. These people are identified with intelligent enrollment and the situation can be thus managed. In addition we also propose a DCT-based feature extractor with threshold coding and discriminability criteria.*

## 1. Introduction

Handwritten signatures have a long tradition in commonly encountered verification tasks like, for example, financial transactions and document authentication. They are normally used and well accepted by the general public, and signatures are easily obtained with relatively cheap devices. These are important advantages of signature recognition over other biometrics. Yet, signature recognition has also some drawbacks: it is a difficult pattern recognition problem due to large possible variations between two signatures of the same person. These variations may be originated by instabilities, emotions, environmental changes, etc, and are personal dependent. In addition, signatures can be easier to forge than other biometrics.

Signature recognition can be split into two categories depending on the data acquisition method:

- Off-line (static), the signature is scanned from a document and the system recognizes the signature analyzing its shape.
- On-line (dynamic), the signature is acquired in real time by a digitizing tablet and the system analyses shape and the dynamics of writing, like for example: positions in the x and y axis, pressure applied by the pen, etc.

Using the dynamic data, further information can be inferred, such as accelerations, velocity, curvature radius, etc.[1]. In this paper, we will focus on this approach.

For a signature verification system, depending on testing conditions and environment, three types of forgeries can be established .
[2]:

- Simple forgery, where the forger makes no attempt to simulate or trace a genuine signature.
- Substitution or random forgery, where the forger uses his/her own signature as a forgery.

- Freehand or skilled forgery, where the forger tries and practices imitating as closely as possible the static and dynamic information of the signature to be forged.

From the point of view of security, the last one is the most damaging; for this reason, some databases suitable for system development include some trained forgeries [3,**4**].

A comprehensive survey of signature verification can be found in [1,.

[2[5]. The existing methods differ both in feature selection and in decision techniques and they can be divided into two classes [6]:

- Feature-based, in which a set of global features is derived from the signature trajectories. Some authors take measures such as total duration of the signature or the number of pen ups, etc. [7[8]. Others compute the Fourier descriptors of the signature trajectory [9[10[11]. Once the features obtained, the decision method is some kind of distance [12], a statistical model [8], or a neural network [13].

- Function-based, in which the time sequences describing the local properties of the signature are used for recognition [14,[**15**],[**16**]. The most usual ones are position, velocity and acceleration of the trajectory, but also the pressure and the pen inclination [17]. Time sequences are then matched by using elastic distance measures such as Dynamic Time Warping [18[19[20] or converted into a sequence of vectors and modeled with Hidden Markov Models [21[22[6].

There are also systems that combine the confidences provided by the two approaches [22[8] and/or use user-dependent decisions, with score normalization techniques [37] or user-dependent thresholds [15], and yield better verification results.

The evaluation of a verification system requires the analysis of two types of errors [24]: Type I error or False Reject Rate (FRR), which is the percentage of genuine signatures incorrectly rejected by the system. Type II error or False Acceptance Rate (FAR), which is the percentage of forgeries incorrectly accepted by the system. These two types of errors usually have different associated costs depending on security requirements. The performance of a system may be measured by the value of the Detection Cost Function (DCF) defined as [25]:

$$DCF = C_{FR} \cdot FRR \cdot P(C) + C_{FA} \cdot FAR \cdot P(I) \qquad (1)$$

where $C_{FR}$ is the cost of a false reject, $C_{FA}$ is the cost of a false acceptance, $P(C)$ is the prior probability of a client (genuine signature) and $P(I)$ is the prior probability of an impostor (forgery). In this paper, in order to facilitate the comparison with other papers, $C_{FR} = C_{FA} = 1$ and $P(C) = P(I) = \frac{1}{2}$ will be assumed. DCF represents the expected cost of making a detection decision based on a weighted sum of False Rejections and False Acceptance error probabilities. A more meaningful performance measure is the error tradeoff curve (DET) [25], which shows how one error changes with respect to the other for several threshold values. Thus, DCF is a scalar value and DET is a plot. DCF value depends on the verification threshold, and it is possible to trade-off the threshold for a minimum DCF (min. DCF).

Another measure of a system's performance, often used in biometrics literature, is the equal error rate (EER) [24], which is the point where the FAR and FRR are the same. In order to facilitate comparison with state-of-the-art performances we have also computed the EER where possible. To obtain the EER value, once the tests have been performed, the threshold is trimmed-up in order to balance FAR and FRR. State-of-the-art EER results for skilled forgeries vary around 6% and 3% [4].

In addition to FAR and FRR there are also two other kind of errors that are usually neglected in laboratory conditions, although they are crucial for real world applications [24]. They are named Failure to Acquire and Failure to Enroll, and are defined as follows:

- Failure to enroll rate (FTE): FTE rate is the expected proportion of the population for whom the system is unable to generate repeatable templates. This will include those unable to present the required biometric feature, those unable to produce an image of sufficient quality at enrolment, and those who cannot reliably match their template in attempts to confirm that the enrolment is usable. The failure to enroll rate will depend on the enrolment policy. For example, in the case of failure, enrolment might be re-attempted at a later date.

- Failure to Acquire rate (FTA): FTA denotes the proportion of times the biometric device fails to capture a sample when the biometric characteristic is presented to it. The failure to acquire rate may depend on adjustable thresholds for signal quality. It is also known as failure to capture.

Normally the literature on biometrics deals with the ideal situation where the set of known users is predefined in advance and no new users are added after the initial set up of the system. In some cases, the problem is even more restricted and it is limited to what is known as closed world problems [23]. In this paper, we go one step further because we deal with the more real situation using two operation steps. In first place, we adjust a biometric signature recognition system by means of a predefined set of initial users. In second place, we study the situation where new users try to enroll in an automatic fashion and the system must be able to detect and manage the Failure to Enroll situations. That is, the system automatically detects training samples which are unstable to create a consistent model and asks the user for a new biometric sample, we call the proposed procedure intelligent enrolment. Experimental results are provided for both phases.

This paper is organized as follows: section two provides an overview of the initial system setup. Section three explains the system operation. Section four presents the experimental results using the MCYT database and section five summarizes the main conclusions.

## *2. Initial system design*

In this section we will discuss the initial set up for the on-line signature verification system, considering a database of genuine and impostor users. Usually, a pattern recognition system consists of two main blocks: feature extraction and classifier. Figure 1 summarizes this scheme for our on-line signature recognition problem. Given a set of measurements provided by a digitizing tablet, we try to obtain the most relevant features for classification and then a classifier is used to compare with a reference.

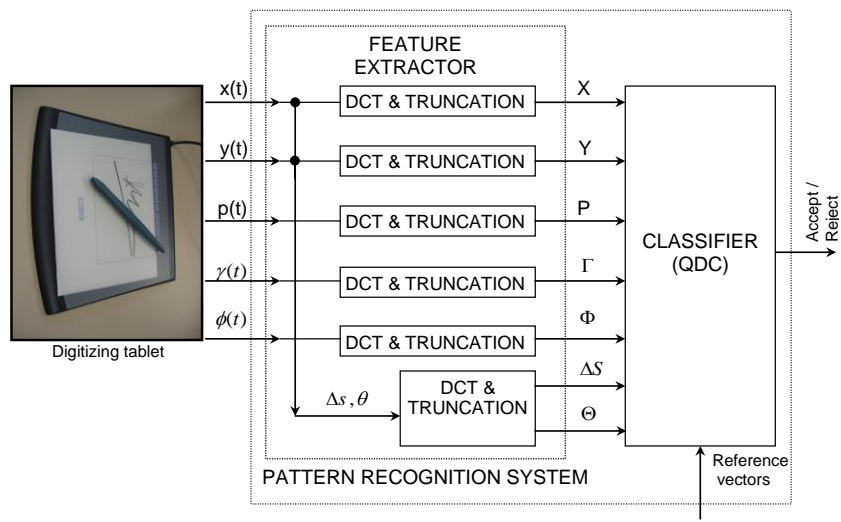

**Figure 1: Pattern recognition system scheme**

## 2.1. Feature extraction

We use a spectral method [9[10,[**11**], but replacing the usual Fast Fourier Transform by the Discrete Cosine Transform (DCT) to perform a low-frequency filtering of temporal variables. The same approach has been previously used for face recognition [26] with spatial variables. This method characterizes the dynamic and global behavior of the signature by taking some of the first terms of the DCT, and does not require preprocessing (re-sampling and smoothing) as other commonly used methods do [15,[**27**]].

Another difference with the usual spectral methods (including [28]) is that, with the help of the classifier (see next section), we perform a threshold coding with discriminability criterion and not the usual zonal or threshold coding with representatibility criterion, commonly used for image compression [29].

The digitizing tablet provides the horizontal and vertical position of the pen over time in equally spaced intervals: $x(n)$ and $y(n)$, $n = 1, ..., N_s$. It also provides the pressure, $p(n)$, the azimuth angle, $\gamma(n)$, and the altitude angle, $\phi(n)$, of the pen at the same intervals [3]. $N_s$ stands for signature time durations in time samples. We complete this five discrete time functions with the displacement: $\delta s(n) = \sqrt{(\delta x)^2 + (\delta y)^2}$, $n = 1,...,N_s - 1$; and the curvature angle: $\theta(n) = \delta \alpha$, $n = 1,...,N_s - 2$; between the intervals. $\delta x$ and $\delta y$ are the coordinate differences between two consecutive points and $\delta \alpha$ are the angle differences between two consecutive points of $\alpha(n) = \arctan \frac{\delta y}{\delta x}$, $n = 1,...,N_s - 1$. The feature extraction performs the DCT of each variable and initially takes the first 30 terms (this number will be explained later), representing a total number of 210 features per signature (7x30). Nevertheless, we do not finally pick up all these 30 coefficients, because we will perform a selection using a discriminability criterion, as described in the next section.

## 2.2. Classifier

In conventional statistical pattern recognition, objects are assigned to classes based on a set of measures extracted from the objects in the form of feature vectors. Usually there are a limited number of classes and a great number of training samples, for example in the recognition of handwritten digits (10 classes) from U.S. postal envelopes described in [30]. This situation is clearly reversed in biometrics, where one normally takes three to five measures per person during the enrollment (three to five samples per class for training) and the enrolled population in the database is large (much more classes than samples per class) [24]. So, the statistical properties of each class are not well defined due to the small number of samples.

In order to overcome this problem, we train a Quadratic Discriminant Classifier (QDC) [31] to solve the dichotomy: *Are these two signatures of the same person?* In doing this, we solve the problem concerning the number of training samples per class and, because we do not need to train the QDC with the people present in the operational database, it will be capable of classifying in an open world situation [32]. The difference between the signature features by pairs is performed. We compute the covariance matrix $\Sigma_C$ (C for clients) with pairs of genuine signatures and the covariance matrix $\Sigma_I$ (I for impostors) with pairs corresponding to a genuine signature and a forgery. Due to the symmetry of the problem (as we can establish the difference between signatures in any order, distributions must be symmetric around the origin) we have $\mu_C = \mu_I = 0$ for the corresponding means. The discriminant function that takes positive values for genuine signature pairs takes the form [31, [**33**]]:

$$g(\vec{x}) = \frac{1}{2}\vec{x}^T\left(\Sigma_I^{-1} - \Sigma_C^{-1}\right)\vec{x} + \frac{1}{2}\ln\frac{|\Sigma_I|}{|\Sigma_C|} + \ln\frac{P(C)}{P(I)} \tag{2}$$

The discriminative power of each feature may be evaluated comparing the corresponding diagonal terms of $\Sigma_C$ and $\Sigma_I$ or their square roots (standard deviations). Let $\sigma_C$ and $\sigma_I$ be the standard deviations for clients and impostors of one feature. Deviations between genuine signatures are expected to be less important than the deviations for forgeries. Then, fixing a threshold for the quotient $\frac{\sigma_C}{\sigma_I}$ is a good and easy feature selection criteria as we will demonstrate with the experimental results (see section four).

Once the features are selected, fixing the $\frac{\sigma_C}{\sigma_I}$ threshold, the QDC is trained by computing the covariance matrices, $\Sigma_C$ and $\Sigma_I$, from the training set of measures. By applying the same set to the matcher we can adjust an initial threshold according to a certain criterion, for example a fixed FRR, a fixed FAR or the one that minimizes the DCF. This initial threshold (T0) is used in the next adjustment stage to determine the operation thresholds of the system.

### 2.3. System adjustment

In order to adjust the operational thresholds of the system we use another set of measures named validation set. It is composed by genuine signatures and forgeries of persons that have not participated in the training set. First we must decide the number of reference signatures to be used for each person enrolled to the system, usually three to five signatures. After capturing and extracting the features of these signatures we establish their consistency, matching them pair wise. In this article for example, we reject a signature out of three references if it mismatches with the other two. We use the initial threshold, T0, obtained in the preceding section. If one signature is rejected it must be substituted by another one from the same person.

We can operate the system with a common threshold (CT) or with individual thresholds (IT), as described next.

**Common threshold operation**

Once the enrollment is finished, we use the remaining genuine and forgery signatures for determining the operational thresholds that fulfill a fixed criterion [24], for example a fixed FRR, a fixed FAR or the one that minimizes the DCF (as we have done in this work).

By matching with only one reference, we obtain the operational threshold for enrollment (TE). This threshold is used during the enrollment to compare the captured references pair wise for discarding outliers.

By fusing the scores provided by the references, we can fix the common operational threshold for verification (CT), obtaining the threshold that minimizes the DCF in our case. In this work, we have tested two fusion methods [34]: mean and maximum of the scores. The best results have been obtained with the maximum, as in [15][20].

**Individual threshold operation**

To adapt the verification process to person-dependent variations explained in the first paragraph of the introduction, we have investigated a bi-parametric individual threshold determination based on the adapted decision explained in [35].

Two types of persons are considered:

- Type 1: Those with the minimum matching score of the enrollment signatures lying under the common threshold, CT. For these individuals we expect that lowering the threshold may have a much more effect in reducing the FRR than in increasing the FAR.

- Type 2: Those with the minimum matching score of the enrollment signatures lying up to the common threshold, CT. For these individuals we expect that increasing the threshold may have a much more effect in reducing the FAR than in augmenting the FRR.

To cope with the degree of variation for the individual thresholds the following equation is proposed:

$$IT = (1-\alpha) \cdot CT + \alpha \cdot \min(ES) \qquad (3)$$

where IT is the individual threshold, CT the common threshold, ES the set of enrollment matching scores for the individual and $0 \leq \alpha \leq 1$. The parameters $\alpha_1$ and $\alpha_2$ (one for each type of person) must be determined during the system adjustment. We have tested the use of the mean or the total variation instead of the minimum function, but the best results have been obtained by this rule. In this work, after adjusting the CT, already explained, tests are made with different values of the parameters (0, 0.1, 0.2… 1) and we take the ones that produce the best results.

We can also compensate for individual variability with user dependent score normalization techniques [36][37]. This step is crucial when combining different matchers in a multibiometric system [38], but it is not necessary in our work.

## 3. System operation

Once the system has been adjusted, it is ready for operation. The system will work with fixed parameters determined by the training and adjustment processes described in section two:

- Features selected by the $\frac{\sigma_C}{\sigma_I}$ threshold.

- Enrollment threshold, TE.

- Common threshold, CT.

- Parameters that determine the individual thresholds, $\alpha_1$ and $\alpha_2$.

The system operation after system adjustment is described as follows.

## 3.1. User enrollment

For the enrollment of the system clients we use the same procedure explained for system adjustment (see the first paragraph of section 2.3), but with the new enrollment threshold, TE. If one of the reference signatures does not achieve the consistency criteria with the others, the system demands an additional signature sample to the user to replace it. With this procedure we can also help the user to deal with the capture process of his/her signature. We can speak of intelligent enrollment because we detect and manage the failure to enroll situations. This is especially important because we avoid the introduction of inconsistent user data which would produce a poor performance of the biometric system.

The people who fail to enroll cannot be managed by the system and we will need to use human abilities or other biometrics to apply with them. This situation is not new in biometrics (for example ~4% of people cannot be enrolled in a fingerprint system [39]) and we can manage this annoyance if we know who they are.

## 3.2. User verification

In the verification process the user claims his/her identity and the system captures his/her signature. The system matches the captured signature with the references stored during the enrollment of the client (claimed identity), obtaining one score for each reference. The system uses the selected fusion technique and performs an *Accept/Reject* decision comparing the resulting score with the CT or the IT depending on the operation mode.

## 4. Experiments and results

### 4.1. Database

We have used our database MCYT [3], acquired with a WACOM graphics tablet. The sampling frequency is set to 100 Hz, obtaining, among other information not used in this paper, the following set for each sampling instant:

- Position in x-axis, $x(t)$: [0-12700], corresponding to 0-127 mm.
- Position in y-axis, $y(t)$: [0-9700], corresponding to 0-97 mm.
- Pressure applied by the pen, $p(t)$: [0-1024].
- Azimuth angle of the pen with respect to the tablet, $\gamma(t)$: [0-3600], corresponding to 0-360 degrees.
- Altitude angle of the pen with respect to the tablet, $\phi(t)$: [300-900], corresponding to 30-90 degrees.

The database consists of 330 people with 25 genuine signatures and 25 skilled forgeries per person. These skilled forgeries are produced by five users after observing the images of the signature to imitate, testing to copy them – almost 10 times – and then, producing the valid forgeries in an easy way (i.e. each individual acting as a forger is requested to sign naturally, without artifacts, such as breaks or slowdowns). We displaced the origin of coordinates to the first point of the signature and normalized spatial distances according to the vertical range of the signature. We do not need a more sophisticated normalization method such as the center of mass, because we will perform the DCT of the signature and we will look for relative frequencies.

For our experiments users, signatures and forgeries have been randomly reordered in order to mitigate order correlations due to session capture and sub-corpus of the MCYT (see [3]). We split the database into three parts:

- DB1, consisting of 80 people and used for training the QDC.

- DB2, consisting of 80 people and used as a validation set for system adjustment.

- DB3, the rest of the database (170 persons) used for testing the system simulating their operation.

The experiments have been performed with three and five references per person, with common and individual thresholds and with intelligent and normal enrollment (without verifying signature consistence). For intelligent enrollment a maximum of 6 (for three references) and 10 (for five references) samples had been used to test the enrollment of a person, if it has not been possible, a Failure to Enroll (FE) is produced. For three references this implies a total of 170x19 genuine tests plus 170x25 impostor tests. For five references we have 170x15 genuine tests plus 170x25 impostor tests.

### 4.2. Classifier training and feature selection

During the training of the QDC with DB1 the feature selection is also produced. The number of selected features depends on the $\frac{\sigma_C}{\sigma_I}$ threshold considered (see Table 1). The features selected from the DCT of each variable are normally taken in natural order, the threshold actuates as a parameter defining the number of terms accepted by the low-frequency filter. We can see that 30 terms per variable are sufficient for our purposes.

| threshold | Features selected from the DCT of | | | | | | | Total |
|---|---|---|---|---|---|---|---|---|
| | $x(n)$ | $y(n)$ | $p(n)$ | $\gamma(n)$ | $\phi(n)$ | $\delta s(n)$ | $\theta(n)$ | |
| 0.35 | 2 | 0 | 1 | 1 | 1 | 0 | 0 | 5 |
| 0.40 | 4 | 2 | 3 | 1 | 1 | 1 | 1 | 13 |
| 0.45 | 7 | 8 | 7 | 1 | 1 | 1 | 2 | 27 |
| 0.50 | 10 | 9 | 9 | 1 | 1 | 4 | 3 | 37 |
| 0.55 | 12 | 11 | 9 | 1 | 1 | 7 | 16 | 57 |
| 0.60 | 15 | 14 | 13 | 1 | 2 | 8 | 27 | 80 |
| 0.65 | 16 | 16 | 18 | 3 | 3 | 12 | 30 | 98 |
| 0.70 | 20 | 16 | 20 | 11 | 20 | 18 | 30 | 135 |

**Table 1: Features selected from the DCT for each variable and total number of features selected for different values of the $\frac{\sigma_C}{\sigma_I}$ threshold. Results obtained in the training phase of the QDC with DB1.**

In Table 2 we see that the performance of the matcher is affected by the number of selected features. Initially the performance increases with increasing $\frac{\sigma_C}{\sigma_I}$ thresholds, because the new features provide more information about the signatures. These new features also contribute with certain noise that masks part of the new information, a maximum of performance is achieved and the performance vanishes with increasing thresholds. The performances shown in Table 2 are optimistic due to the fact that they have been measured with the same set used to train the classifier. In the next section we use the validation set, DB2, to obtain better predictions of performance and adjustment thresholds.

Although pen inclination trajectories have shown some discriminative capabilities in other works [40][41], our results are in line with recent research [6] showing that pen inclination information is not as discriminative as the trajectory information in dynamic signature; only two coefficients are present near the threshold of maximum discriminability (~0.5). The features extracted from the position and the pressure applied by the pen represent 28 of the 37 total features.

| $\frac{\sigma_C}{\sigma_I}$ threshold | min. DCF |
|---|---|
| 0.35 | 10.77 |
| 0.40 | 9.58 |
| 0.45 | 9.27 |
| 0.50 | 9.28 |
| 0.55 | 10.49 |
| 0.60 | 11.39 |
| 0.65 | 11.24 |
| 0.70 | 11.08 |

**Table 2: Performance of the QDC, measured with DCF in %, trained on DB1 for the pairs of genuine-genuine and genuine-forgery signatures of the same training set.**

### 4.3. System adjustment

The system has been adjusted as explained in section 2.3, with the validation set DB2. We have chosen to optimize the DCF as the criteria for adjusting the thresholds. The maximum fusion method has obtained the best results, with $\frac{\sigma_C}{\sigma_I} = 0.45$ as the best threshold for feature selection.

Table 3 shows the adjustment parameters with the minimum DCF obtained for values of the $\frac{\sigma_C}{\sigma_I}$ threshold near the best one. It can be observed that the performances for $\frac{\sigma_C}{\sigma_I} = 0.45$ and $\frac{\sigma_C}{\sigma_I} = 0.50$ are very close, the difference is not statistically significant.

| $\frac{\sigma_C}{\sigma_I}$ threshold | 3 References | | | | | 5 References | | | | |
|---|---|---|---|---|---|---|---|---|---|---|
| | TE | CT | $\alpha_1$ | $\alpha_2$ | min. DCF | TE | CT | $\alpha_1$ | $\alpha_2$ | min. DCF |
| 0.40 | 17.10 | 20.57 | 0.2 | 0.8 | 8.75 | 17.16 | 20.72 | 0.1 | 0.9 | 7.43 |
| 0.45 | 28.78 | 34.18 | 0.1 | 0.7 | 7.45 | 29.16 | 36.82 | 0.6 | 0.9 | 6.52 |
| 0.50 | 35.27 | 44.69 | 0.1 | 0.5 | 7.85 | 37.75 | 46.39 | 0.5 | 0.8 | 6.58 |
| 0.55 | 43.13 | 58.52 | 0.6 | 0.7 | 8.33 | 49.35 | 62.52 | 0.5 | 0.8 | 7.08 |

**Table 3: Adjustment parameters and performance, measured with DCF in %, obtained for the validation set, DB2. The performance is measured with the individual threshold operation and the maximum fusion method is used.**

Once obtained the parameters of the system we can deal with system operation.

### 4.4. System operation performance

In this section we have tested the system by using DB3 and simulating operational conditions. We have enrolled every user of DB3. The consistency of the enrollment signatures is established by matching them pair wise using the threshold TE; for three references a signature is rejected if it mismatches with the other two and, for five references, if it mismatches with three or more. A maximum of 6 (for three references) and 10 (for five references) samples had been used to test the enrollment of a person; if it has not been possible a Failure to Enroll (FE) is produced.

After enrollment we have tested the performance of the system with the genuine and impostor tests, and using the predefined parameters CT, $\alpha_1$ and $\alpha_2$ (a-priori results). We have also computed the a-posteriori results, computing new values for the parameters in order to minimize the DCF. Comparing both results we can assess the robustness of the system, in the sense that the adjustment done gives good (robust) or poor results during operation.

We must mention that the a-priori results presented in this work do not only fix the parameters that define the individual threshold ($\alpha_1$ and $\alpha_2$ in our case) as is the case of other works [37],[**8**], but we also fix the decision threshold in order to demonstrate the practical applicability like in a real operation condition.

For the two types of operation modes the following results have been achieved.

**Common threshold**

Table 4 shows the performance of the proposed system. We have obtained these results with intelligent enrollment, common threshold and using the maximum fusion method. The a-priori DCF is calculated using the system in normal operation, with the CT obtained in the adjustment process. The min. DCF is calculated a-posteriori, computing a new threshold.

| $\frac{\sigma_C}{\sigma_I}$ threshold | 3 References | | 5 References | |
|---|---|---|---|---|
| | DCF a-priori | min. DCF | DCF a-priori | min. DCF |
| 0.35 | 10.67 | 10.48 | 9.78 | 9.35 |
| 0.40 | 9.01 | 8.83 | 7.98 | 7.85 |
| 0.45 | 7.89 | 7.86 | 7.18 | 7.10 |
| 0.50 | 8.22 | 7.99 | 7.02 | 6.81 |
| 0.55 | 9.55 | 9.48 | 8.10 | 7.99 |
| 0.60 | 11.03 | 10.85 | 9.28 | 9.23 |
| 0.65 | 11.33 | 10.92 | 9.50 | 9.37 |
| 0.70 | 11.80 | 11.65 | 11.04 | 10.44 |

**Table 4: Performance of the system measured with DCF in % for different values of the $\frac{\sigma_C}{\sigma_I}$ threshold. We have used intelligent enrollment, common threshold and the maximum fusion method. Results obtained with the testing dataset, DB3.**

Figure 2 shows a plot of the performances of the proposed system as they appear in Table 4. The maximum performance (minimum DCF) takes place at the $\frac{\sigma_C}{\sigma_I}$ thresholds of 0.45 (three references) and 0.5 (five references). The results a-priori and a-posteriori are very close, so the proposed system is robust in the sense that the adjustment done gives good results during operation.

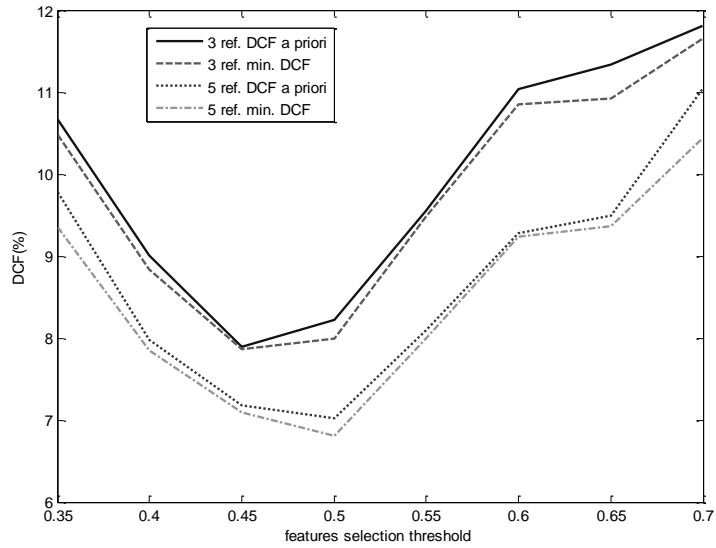

**Figure 2: Plot of the data of Table 4. The feature selection threshold is the $\frac{\sigma_C}{\sigma_I}$ threshold and the min. DCF is computed a-posteriori.**

Our a-posteriori results perform similarly to other state-of-the-art works that use MCYT with five references. They are not directly comparable because the experimental conditions are not the same, but they serve as a reference. With a more complex system like DTW, [20] has obtained a min. DCF of 8.94% and [8], for global information, has obtained an EER of 6.89% (we have computed an EER for five references and $\frac{\sigma_C}{\sigma_I} = 0.50$ of 7.01%).

It is interesting to note that, for five references and $\frac{\sigma_C}{\sigma_I} = 0.50$, the a-priori results correspond to a FRR of 7.26% and a FAR of 6.77%, which is a quite balanced situation.

**Individual threshold**

Table 5 shows the performance of the proposed system. These results have been achieved with intelligent enrollment, individual thresholds and using the maximum fusion method. The a-priori

DCF is calculated using the system in normal operation, with the IT computed from the CT, $\alpha_1$ and $\alpha_2$ obtained in the adjustment process. The min. DCF is calculated a-posteriori, computing a new CT and news $\alpha_1$ and $\alpha_2$.

| threshold | 3 References | | 5 References | |
|---|---|---|---|---|
| | DCF a-priori | min. DCF | DCF a-priori | min. DCF |
| 0.35 | 9.27 | 9.21 | 8.80 | 8.00 |
| 0.40 | 7.94 | 7.12 | 6.41 | 6.15 |
| 0.45 | 7.05 | 6.24 | 5.26 | 5.18 |
| 0.50 | 7.16 | 6.11 | 5.29 | 5.17 |
| 0.55 | 6.87 | 6.75 | 6.09 | 5.97 |
| 0.60 | 7.80 | 7.59 | 6.73 | 6.53 |
| 0.65 | 8.25 | 7.84 | 8.04 | 6.90 |
| 0.70 | 8.50 | 8.53 | 9.07 | 7.50 |

**Table 5: Performance of the system measured with DCF in % for different values of the $\frac{\sigma_C}{\sigma_I}$ threshold. We have used intelligent enrollment, individual threshold and the maximum fusion method. Results obtained with the testing dataset, DB3.**

Figure 3 shows a plot of the performances of the proposed system as they appear in Table 5. The maximum performance (minimum DCF) takes place near the $\frac{\sigma_C}{\sigma_I}$ thresholds of 0.45 and 0.5. The a-priori and a-posteriori results near the points of maximum performance are very similar for five references whereas they differ substantially for three references; this is originated by the small number of references in the individual threshold selection. For five references the system proposed is robust in the sense that the adjustment done gives good results during operation.

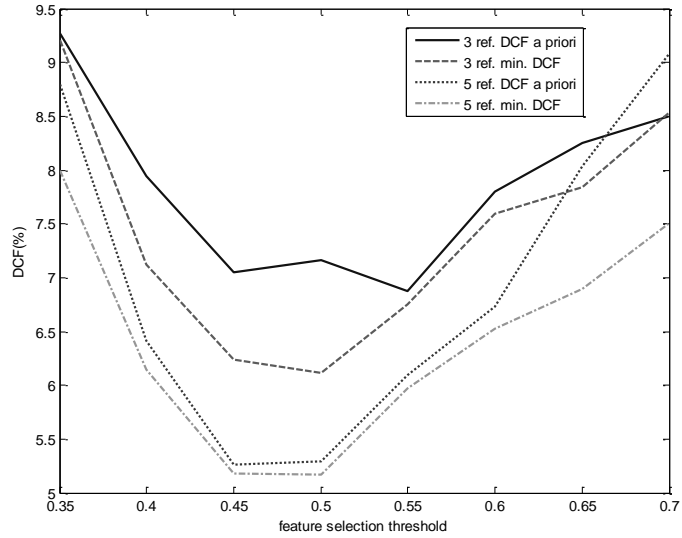

**Figure 3: Plot of the data of Table 5. The feature selection threshold is the $\frac{\sigma_C}{\sigma_I}$ threshold and the min. DCF is computed a-posteriori.**

Compared with the system operating in common threshold at five references and $\frac{\sigma_C}{\sigma_I} = 0.5$, we achieve an increase of performance of almost 25% in both, a-priori and a-posteriori, cases.

In Table 6 we can compare the values for the adjusted parameters (CT, $\alpha_1$ and $\alpha_2$ a-priori) with those computed in the test (a-posteriori) for $\frac{\sigma_C}{\sigma_I} = 0.5$ and five references. We see that the adjusted and tested CT are very close because the computed scores spread between – 3221.00 and 95.14. Figure 4 shows the dependency of the DCF with respect to $\alpha_1$ and $\alpha_2$ for DB2 (adjustment) and DB3 (test).

|  | Adjusted (a-priori) | Tested (a-posteriori) |
|---|---|---|
| CT | 46.39 | 48.02 |
| $\alpha_1$ | 0.5 | 0.5 |

| $\alpha_2$ | 0.8 | 0.6 |

**Table 6: Adjusted with DB2 (a-priori) and tested with DB3 (a-posteriori) values of the parameters of the system at $\frac{\sigma_c}{\sigma_I} = 0.5$, with 5 references and using maximum fusion.**

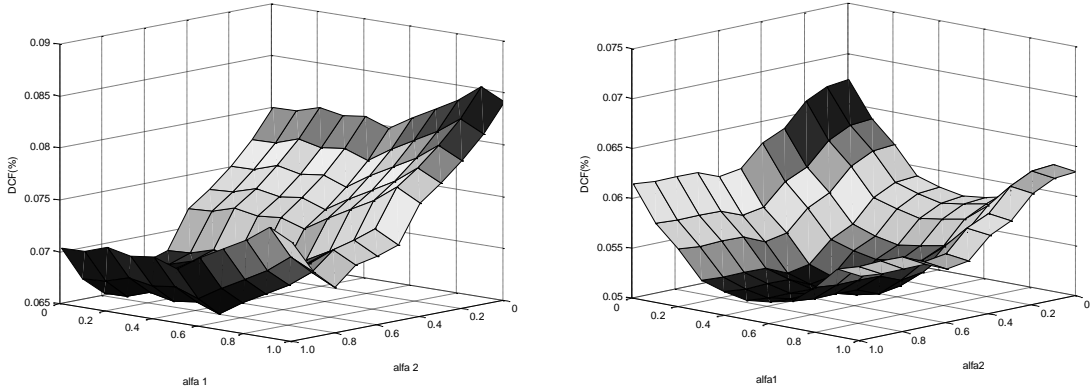

**Figure 4: Left – values of $\alpha_1$ and $\alpha_2$ with the corresponding value of the DCF in % ($\frac{\sigma_c}{\sigma_I} = 0.5$ and 5 references) for DB2 (adjustment). The minimum takes place at $\alpha_1 = 0.5$ and $\alpha_2 = 0.8$. Right – the same values for DB3 (test). The minimum takes place at $\alpha_1 = 0.5$ and $\alpha_2 = 0.6$.**

It is interesting to note that for five references and $\frac{\sigma_C}{\sigma_I} = 0.5$ the a-priori results correspond to a FRR of 5.77% and a FAR of 4.82%, which is also a quite balanced situation.

### 4.5. Effects of intelligent enrollment

Table 7 shows the results of intelligent enrollment for the feature selection threshold $\frac{\sigma_c}{\sigma_I} = 0.5$. Remember that we established the consistency of the enrollment signatures by matching them pair wise. For three references we reject a signature if it mismatches with the other two and, for five references, if it mismatch with three or more. A maximum of 6 (for three references) and 10 (for five references) samples had been used to test the enrollment of a person; if it has not been possible, a failure to enroll (FTE) is produced. Failure to enroll is very important in this system simulation; we expect that in real operation the user will learn, because of system feedbacks, to sign better with the digitizing tablet.

|     | 3 References | | 5 References | |
| --- | --- | --- | --- | --- |
|     | Persons | % | Persons | % |
| + 0 | 135 | 79.41 | 117 | 68.82 |
| + 1 | 17 | 10.00 | 28 | 16.47 |
| + 2 | 7 | 4.12 | 5 | 2.94 |
| + 3 | 1 | 0.59 | 3 | 1.76 |
| + 4 | --- | | 2 | 1.18 |
| + 5 | --- | | 1 | 0.59 |
| FE | 10 | 5.88 | 14 | 8.24 |

**Table 7: Results of intelligent enrollment for the feature selection threshold $\frac{\sigma_C}{\sigma_I} = 0.5$. The numbers with the plus sign represent the number of extra signatures needed for enrollment. FE is the failure to enroll. Results obtained with the testing dataset, DB3, and the maximum fusion method.**

In Table 8 we see a comparison of the performances of the system working with intelligent and normal enrollment at $\frac{\sigma_C}{\sigma_I} = 0.5$ with the maximum fusion method. For five references, intelligent enrollment leaves out 8% of the users, and achieves an increase of performance around 22% using common or individual thresholds, in both a-priori and a-posteriori cases. The individual threshold, as expected, mitigates the negative effects of large variation signers. Intelligent enrollment intends to identify those signers with variation not manageable by the system, and prevents their enrollment. As the same increase in performance for common and individual threshold is obtained, we can see that intelligent enrollment manages part of those people who are not managed by individual threshold. Therefore, individual threshold and intelligent enrollment are complementary.

|  |  | 3 References | | 5 References | |
|---|---|---|---|---|---|
|  |  | DCF a-priori | min. DCF | DCF a-priori | min. DCF |
| Intelligent | IT | 7.16 | 6.11 | 5.29 | 5.17 |
| Intelligent | CT | 8.22 | 7.99 | 7.02 | 6.81 |
| Normal | IT | 7.59 | 6.59 | 6.89 | 6.60 |
| Normal | CT | 9.66 | 9.26 | 9.16 | 8.86 |

**Table 8: Comparison of the performance of the system ($\frac{\sigma_c}{\sigma_I} = 0.5$ and maximum fusion method) with respect to the type of enrollment (intelligent, normal) and threshold (common, individual). The performance is measured with the DCF in %. Results obtained with the testing dataset, DB3.**

Figure 5 shows visual examples of signatures taken from the enrollment process for three references, the first three (top) belong to a user that has been accepted without any new attempt, the second three (middle) to a user that has been accepted with one new attempt (the third signature has been rejected), and the other three (bottom) to a user who has failed to enroll.

We remark that we have worked with a preexisting database, simulating intelligent enrollment without user interaction. Another advantage of intelligent enrollment that must be proved in a real operating system is the user learning from the system feedback. The feedback of the system will help the user to improve their signature abilities with the system interface. These results are encouraging to do future work in this direction.

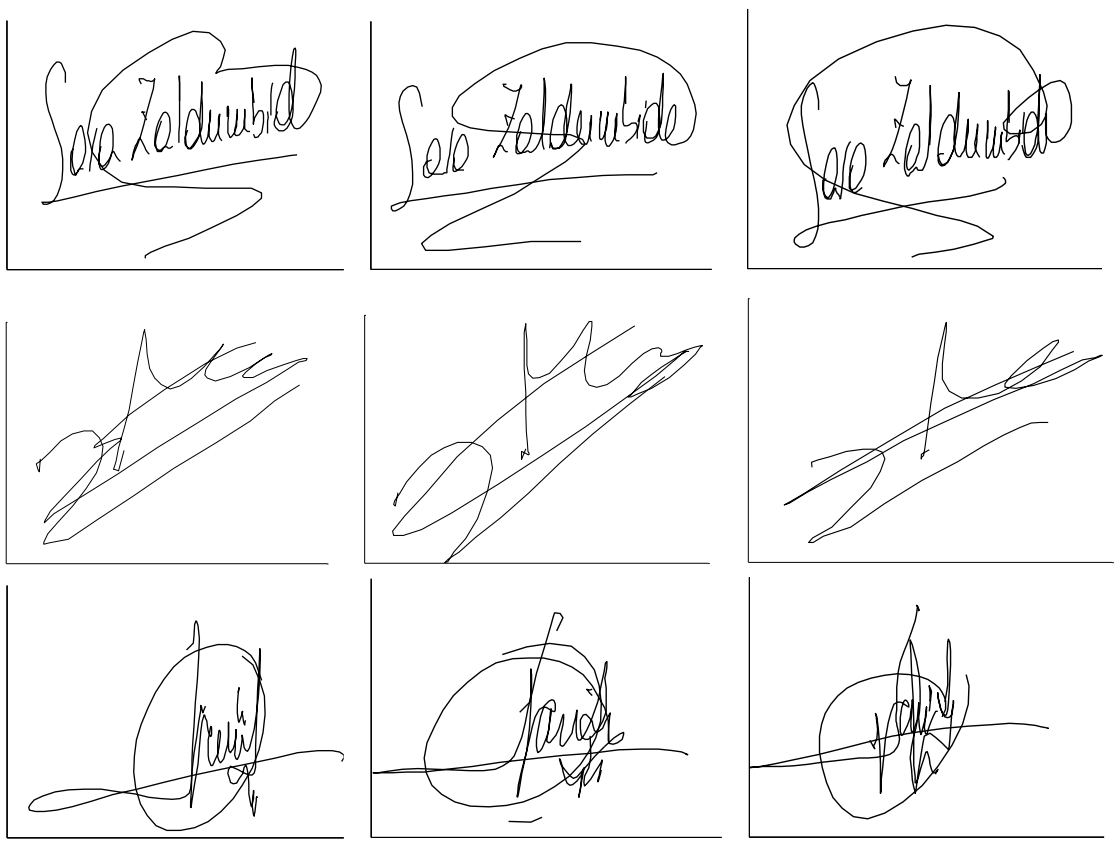

**Figure 5: Visual examples of signatures belonging to users that had been accepted without any new attempt (top), accepted with one new attempt (middle), and not accepted (bottom).**

### 4.6. Comparison between mean and maximum fusion

The maximum fusion always outperforms the use of the mean fusion in our work, as an example we present in Table 9 the results at $\frac{\sigma_C}{\sigma_I} = 0.5$ for the mean method, for comparison with the same terms of Table 4 and Table 5. In Figure 6 we can compare the DET curves corresponding to the two fusion methods for 5 references and common threshold.

|    | 3 References |          | 5 References |          |
|----|--------------|----------|--------------|----------|
|    | DCF a-priori | min. DCF | DCF a-priori | min. DCF |
| IT | 7.45         | 7.27     | 5.77         | 5.73     |
| CT | 9.80         | 9.20     | 7.99         | 7.87     |

**Table 9: Performance of the system ($\frac{\sigma_c}{\sigma_I} = 0.5$) for the mean fusion method with respect to the type of threshold (common, individual). The performance is measured with the DCF in %. Results obtained with the testing dataset, DB3.**

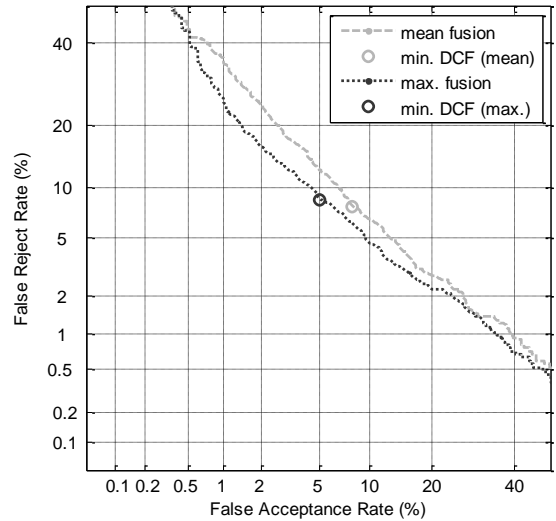

**Figure 6: DET curves for $\frac{\sigma_c}{\sigma_I} = 0.5$ and 5 references with common threshold. The curves correspond to the mean and maximum fusion methods. Results obtained with the testing dataset, DB3.**

### 4.7. Performance of the system for random forgeries

The system has been trained with skilled forgeries; in this section we study the performance of the system with random forgeries using one authentic signature from all the other enrolled writers as random forgeries. There are a total number of 170x169 random forgeries.

**Common threshold**

Table 10 shows the performance measured by the DCF expressed in %. Like in section 4.2, these results are obtained with intelligent enrollment and using the maximum fusion method. The a-priori DCF is calculated by using the system in normal operation, with the CT obtained in the adjustment process. The min. DCF is calculated a-posteriori, by computing a new CT.

| threshold | 3 References | | 5 References | |
|---|---|---|---|---|
| | DCF a-priori | min. DCF | DCF a-priori | min. DCF |
| 0.35 | 7.10 | 5.64 | 5.63 | 5.08 |
| 0.40 | 5.06 | 3.82 | 3.74 | 3.11 |
| 0.45 | 3.84 | 2.71 | 3.74 | 2.24 |
| 0.50 | 4.84 | 2.87 | 3.74 | 2.16 |
| 0.55 | 4.83 | 3.35 | 4.15 | 2.61 |
| 0.60 | 5.46 | 4.08 | 4.78 | 3.14 |
| 0.65 | 6.44 | 4.08 | 4.95 | 3.47 |
| 0.70 | 5.53 | 4.89 | 6.37 | 4.12 |

**Table 10: Performance of the system for random forgeries measured with DCF in % for different values of the $\frac{\sigma_C}{\sigma_I}$ threshold. We have used intelligent enrollment, common threshold and the maximum fusion method. Results obtained with the testing dataset, DB3.**

Figure 7 shows a plot of performances as they appear in Table 10. The maximum performance (minimum DCF) takes place for $\frac{\sigma_C}{\sigma_I}$ near 0.45 and 0.5. The results a-priori and a-posteriori are not as close as for skilled forgeries; this suggests that training and adjusting the system only with skilled forgeries seems to lower the performance for random forgeries. Nevertheless we still achieve good error values, below 4%. Future work must be done to study how to overcome this effect without lowering the capacity of the system with skilled forgeries.

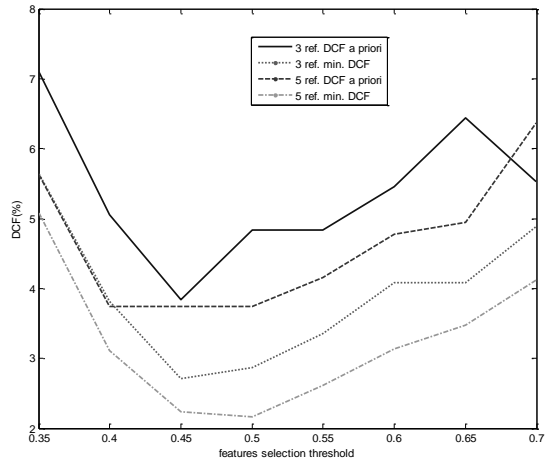

**Figure 7: Plot of the data of Table 10. The feature selection threshold is the $\frac{\sigma_C}{\sigma_I}$ threshold and the min. DCF is computed a-posteriori.**

**Individual threshold**

Table 11 shows the performance measured by the DCF expressed in %. Like in section 4.2, these results are obtained with intelligent enrollment and using the maximum fusion method. The a-priori DCF is calculated by using the system in normal operation, with the IT computed from the CT, $\alpha_1$ and $\alpha_2$ obtained in the adjustment process. The min. DCF is calculated a-posteriori, computing a new CT and new $\alpha_1$ and $\alpha_2$.

| threshold | 3 References | | 5 References | |
|---|---|---|---|---|
| | DCF a-priori | min. DCF | DCF a-priori | min. DCF |
| 0.35 | 6.39 | 5.41 | 4.74 | 4.70 |
| 0.40 | 5.87 | 3.25 | 3.45 | 2.27 |
| 0.45 | 4.97 | 2.31 | 2.61 | 1.83 |
| 0.50 | 5.09 | 2.40 | 2.99 | 1.89 |
| 0.55 | 4.58 | 2.57 | 3.27 | 2.38 |
| 0.60 | 5.34 | 3.23 | 3.81 | 2.69 |
| 0.65 | 5.40 | 3.44 | 5.49 | 2.80 |
| 0.70 | 5.83 | 4.14 | 6.13 | 3.22 |

**Table 11: Performance of the system for random forgeries measured with DCF in % for different values of the $\frac{\sigma_C}{\sigma_I}$ threshold. We have used intelligent enrollment, individual threshold and the maximum fusion method. Results obtained with the testing dataset, DB3.**

Figure 8 shows a plot of performances as they appear in Table 11. The maximum performance (minimum DCF) takes place near the $\frac{\sigma_C}{\sigma_I}$ thresholds of 0.45 and 0.5. The results a-priori and a-posteriori near the points of maximum performance are similar for five references whereas they differ substantially for three references; this is caused by the small number of references in the individual threshold selection. For five references the system working with individual thresholds is more robust than for common threshold.

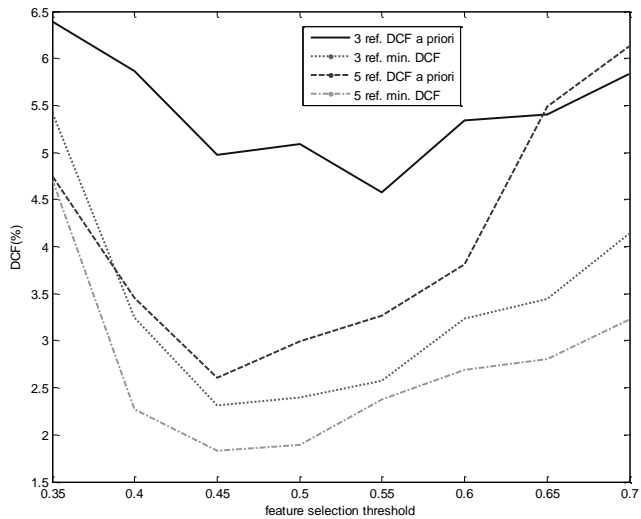

**Figure 8: Plot of the data of Table 11. The feature selection threshold is the $\frac{\sigma_C}{\sigma_I}$ threshold and the min. DCF is computed a-posteriori.**

With the introduction of the individual threshold we achieve an increase of performance of 20% in the a-priori case and of 12.5% in the a-posteriori case for five references and $\frac{\sigma_C}{\sigma_I} = 0.5$. These results are in accordance with state-of-the-art methods [4].

## *5. Conclusions*

In this paper a new system for on-line signature verification has been presented. We have performed a simulation with the largest available on-line signature database, MCYT, which consists of 330 people with genuine and skilled forgeries performed by 5 other different users. The experimentation, although performed in a laboratory, has been developed keeping in mind the casuistic of real world, putting special emphasis on the following problems:

a) Open world situation with failure to enroll detection and managing.

b) Verification errors obtained with a decision threshold which has been computed a-priori.

The main contribution of this work is the management of the failure to enroll situations by means of a new proposal, called intelligent enrollment, which consists of consistency checking in order to automatically reject low quality samples. This strategy lets to enhance the verification errors up to 22%, when leaving out 8% of the users, using common or individual thresholds, in both a-priori and a-posteriori cases. Individual threshold and intelligent enrollment are complementary. Individual threshold mitigates the negative effects of large variation signers. Intelligent enrollment identifies signers with variation which are not manageable by the system, and prevents their enrollment. In this situation, 8% of the people cannot be enrolled in the system and must be verified by other biometrics or by human abilities. These persons are identified with intelligent enrollment and the situation can thus be managed. This situation is not new in biometrics, ~4% of people cannot be enrolled in a fingerprint system due to the poor quality of their fingerprints. In a signature verification system, due to the large variability among signatures of the same person, this percentage is expected to be greater. We also expect that some people may be trained by the system feedback during intelligent enrollment to overcome this problem. This work is a first contribution to this important practical topic in signature verification.

We also provide verification errors with a-priori fixed decision parameters (including the decision threshold), with the system adjusted by means of a validation set. This is another novelty because presently the signature verification literature does not include such a-priori results, necessary to verify the stability of the system and its practical capabilities. In this most real and worst situation, far from the laboratory condition, we achieve a 5.26% minimum detection cost function (skilled forgeries) which is comparable to the state-of-the-art performances.

Finally, a novel spectral method that uses the discrete cosine transform, with a threshold coding with representatibility criterion, is employed for feature extraction.

**ACKNOWLEDGEMENTS**

This work has been supported by FEDER and MEC, TEC2006-13141-C03-02/TCM

35